\documentclass[onecolomn, journal]{IEEEtran}

\pdfoutput=1
\usepackage{multirow}
\usepackage{amsmath}
\usepackage{graphicx}
\usepackage{epstopdf}
\usepackage{amsfonts}
\usepackage{subfigure}
\usepackage[colorlinks,linkcolor=black,anchorcolor=black,citecolor=black]{hyperref}
\usepackage{cite}
\usepackage{makecell}
\usepackage{booktabs}
\usepackage{amssymb}
\usepackage{algorithm,algorithmic}
\usepackage{setspace}
\usepackage{array}
\usepackage{colortbl}
\usepackage{xcolor}

\begin{document}

\title{Dig-CSI: A Distributed and Generative Model Assisted CSI Feedback Training Framework}

\author{Zhilin~Du,
        Haozhen~Li,~\IEEEmembership{Graduate Student Member,~IEEE},
        Zhenyu~Liu,~\IEEEmembership{Member,~IEEE},
        Shilong~Fan,
        Xinyu~Gu,~\IEEEmembership{Member,~IEEE}, 
        and~Lin~Zhang,~\IEEEmembership{Member,~IEEE}
\thanks{This work was supported in part by the National
Key Research and Development Program under Grant 2022YFF0610303, and in part by the National Natural Science Foundation of China (NSFC) under Grant 62201089.}
\thanks{Z. Du, H. Li, Z. Liu, S.Fan, L. Zhang, X. Gu are with the School of Artificial Intelligence, Beijing University of Posts and Telecommunications, Beijing, X. Gu is also with the Purple Mountain Laboratories, Nanjing 211111, China.
100876, China (e-mail: \{duzhilin, lihaozhen, lzyu, fanshilong, guxinyu, zhanglin\}@bupt.edu.cn).
}
}

\maketitle

\begin{abstract}

The advent of deep learning (DL)-based models has significantly advanced Channel State Information (CSI) feedback mechanisms in wireless communication systems. However, traditional approaches often suffer from high communication overhead and potential privacy risks due to the centralized nature of CSI data processing. To address these challenges, we design a CSI feedback training framework called Dig-CSI, in which the dataset for training the CSI feedback model is produced by the distributed generators uploaded by each user equipment (UE), but not through local data upload. Each UE trains an autoencoder, where the decoder is considered as the distributed generator, with local data to gain reconstruction accuracy and the ability to generate. Experimental results show that Dig-CSI can train a global CSI feedback model with comparable performance to the model trained with classical centralized learning with a much lighter communication overhead.

\end{abstract}


\IEEEpeerreviewmaketitle

\section{Introduction}

\IEEEPARstart{I}{n} a massive multiple-input multiple-output system (MIMO) with frequency division duplex (FDD) mode, the base station (BS) requires channel state information (CSI) from user equipment (UE). The transmission of CSI (namely CSI feedback task) includes compression and reconstruction of the CSI matrix to save the wireless channel source. Deep learning (DL) based CSI feedback has achieved remarkable success with many deep neural network (DNN) models proposed \cite{CsiNet}\cite{CRNet}\cite{CLNet}. These models can obtain high reconstruction accuracy of CSI matrices within a certain area. 

\begin{figure*}[tb]
    \centering
    \includegraphics[width=\linewidth]{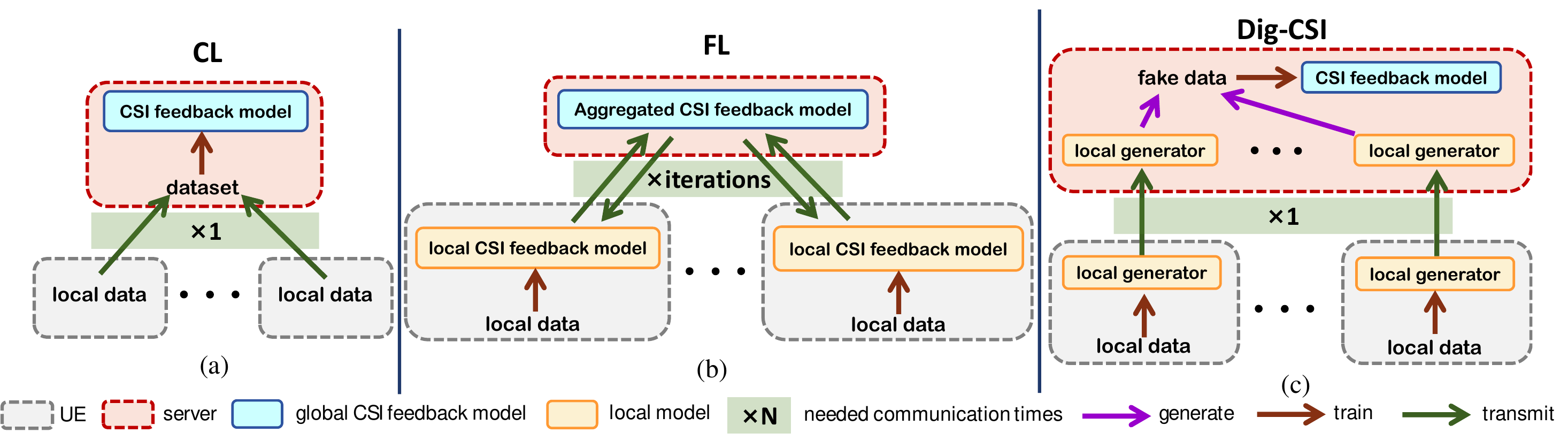}
    \caption{The training framework of CL, FL, and the proposed Dig-CSI.}
    \label{frameworks}
\end{figure*}

In a real deployment, the model is trained at a server with a prepared CSI dataset and is then sent to each UE and BS, which is named centralized learning (CL). The construction of the dataset is crucial in CL, which is either through asking each UE to upload the locally produced CSI matrices as depicted in Fig.\ref{frameworks}(a) or through channel measurement that utilizes specialized equipment to collect CSI matrices. In the former method, the communication overhead between the UE and the server is too heavy and the privacy of the UE is susceptible to leakage while in the latter method, the equipment is expensive and the collection activity is time-consuming and laborious.

In terms of these flaws of CL for DL-based CSI feedback, Federated Learning (FL) is introduced by \cite{FedB} and \cite{FedC} as an improved training method which is depicted in Fig.\ref{frameworks}(b), where each UE trains its model with local data and uploads it to the server and the server aggregates these models to produces a global model for each UE to continue training. Adopting FL for DL-based CSI feedback alleviates UE from uploading local data to the server, which greatly reduces the communication overhead generated in CL and protects the privacy of UE. However, as FL needs several iterations to converge, the overhead of transmitting the model between a UE and the server rises by times as the iteration increases. Moreover, the FL algorithm FedAvg \cite{fedavg} used in \cite{FedB} and \cite{FedC} that merely averages the parameters of each local model encounters a severe problem of client drift \cite{scaffold} that harms the performance of the model when the channel state of each UE obviously differs.

In this paper, we propose a novel training framework that involves \textbf{di}stributed \textbf{g}enerative model for \textbf{CSI} feedback (Dig-CSI) which is depicted in Fig.\ref{frameworks}(c). Dig-CSI absorbs the advantage of FL which uploads local models to save communication overhead and make sure the global model can be trained as in CL through a generative local model, which solves the client drift problem in FL. Additionally, the model in Dig-CSI only needs to be uploaded once, therefore significantly reducing the communication overhead in FL. We use the structure of autoencoder to implement the local model and adopt sliced Wasserstein distance introduced in \cite{SWAE} to train the distributed generator. We set a scenario where multiple UEs participate in CSI feedback training and deploy CL, FL, and Dig-CSI on it. Experimental results show that Dig-CSI achieves the least communication overhead and has a comparable performance as CL.

\section{System model}
\label{sec 2}
Supposing that in an FDD Massive MIMO system, a BS serving for multiple single-antenna UEs is equipped with an $N_t$-antenna uniform linear array (ULA). Considering the use of orthogonal frequency division multiplexing (OFDM) over $N_f$ subcarriers, the received signal on the $n^{th}$ subcarrier for a UE can be modeled as:

\begin{equation}
    y_n=\mathbf{h}_n^H\mathbf{v}_nx_n+z_n
\end{equation}
where $\mathbf{h}_n\in\mathbb{C}^{N_t}$, $\mathbf{v}_n\in\mathbb{C}^{N_t}$, $x_n\in\mathbb{C}$ and $z_n\in\mathbb{C}$ denote the channel vector, the precoding vector, the transmit data symbol and the additive noise, respectively. The downlink CSI matrix held at the UE can be expressed as:

\begin{equation}
    \mathbf{H}=[\mathbf{h}_1,\mathbf{h}_2,\cdots, \mathbf{h}_{N_f}]^H\in\mathbb{C}^{N_f\times N_t}
\end{equation}
$\mathbf{H}$ is in antenna-frequency domain, which needs to be converted to angular-delay domain to acquire sparsity. The transformation can be expressed as:
\begin{equation}
    \tilde{\mathbf{H}}=\mathbf{F}_d\mathbf{H}\mathbf{F}_a^H
\end{equation}
where $\mathbf{F}_d\in\mathbb{C}^{N_f\times N_f}$ and $\mathbf{F}_a\in\mathbb{C}^{N_t\times N_t}$ are discrete Fourier transform (DFT) matrices. In DL-based CSI feedback, the UE compresses $\tilde{\mathbf{H}}$ to a codeword $\mathbf{s}$ with an encoder, which can be expressed as 
\begin{equation}
    \mathbf{s} = f_{en}(\tilde{\mathbf{H}}; \Theta_{en}),
\end{equation}
where $\Theta_{en}$ represents the parameters of the encoder. Then $\mathbf{s}$ is sent to the BS through a wireless channel and the BS uses a decoder to reconstruct CSI, which can be expressed as
\begin{equation}
    \hat{\mathbf{H}} = f_{de}(\mathbf{s}; \Theta_{de}),
\end{equation}
where $\hat{\mathbf{H}}$ is the reconstructed CSI, and $\Theta_{de}$ represents the parameters of the decoder.

\section{The Proposed Dig-CSI}
\label{sec 3}
In this section, we first express the framework of Dig-CSI. Then, we show in detail how the generator is implemented, including the training method and the neural network structure.

\subsection{The framework of Dig-CSI}
In this subsection, the framework of our proposed Dig-CSI shown in Fig.\ref{frameworks}(c) is described in detail. Supposing several UE are participating in training and the global model in the server is trained with local datasets collected by these UE. On the UE side, each UE trains its local generator with its local dataset. The well-trained local generator is expected to have the ability to generate data that is similar to the data pieces in the local dataset without any assistance of extra information. Then these local generators are uploaded to the server. On the server side, these local generators produce masses of fake data pieces, which are collected to build a dataset to train the global CSI feedback model. The training of the global CSI feedback model is the same as that in CL.

\subsection{Implementation of Dig-CSI}
In this subsection, we explain how the local generators in Dig-CSI produce data that is similar to local data and express the network structure of local generators.

\subsubsection{The generating data method of the local generator}
\label{sec 3.2.1}

The expected function of the local generator in Dig-CSI corresponds to that of generative models which have been thoroughly studied with many implementations being proposed such as Generative Adversarial Networks (GANs) \cite{GANreview}, Variational AutoEncoders (VAEs) \cite{VAEreview} and so on. Among these, we adopt VAE to design our local generator. Therefore, the structure of each local model is an autoencoder that is made up of an encoder and a decoder where the decoder is the generator that is to be uploaded. The training objective of the local model is to have a better reconstruction accuracy on local data as well as to let the distribution of the latent code output by the encoder close to a pre-defined distribution such as the basic normal distribution. Through this training, the decoder is expected to work independently of the encoder by converting a sample from the pre-defined distribution to a sample that can be considered to belong to the local dataset. Therefore the decoder becomes a local generator.

\subsubsection{Implementation of the local generator}
\label{sec 3.2.2}
\begin{figure*}[tb]
    \centering
    \includegraphics[width=0.9\linewidth, scale=1.00] {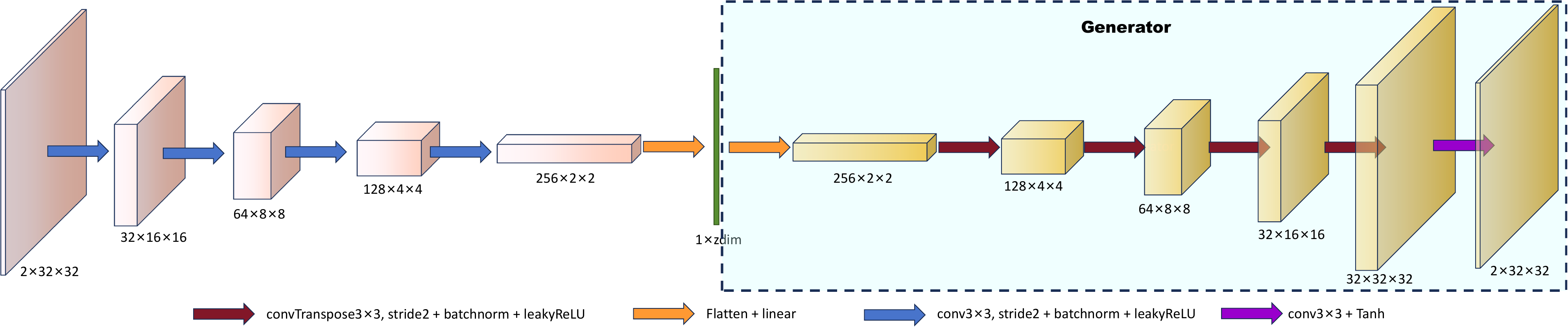}
    \caption{The structure of the local model and local generator in Dig-CSI.}
    \label{generator structure}
\end{figure*}

The local generator needs to accurately rebuild the local dataset, therefore the feedback accuracy during training of the local model is crucial. However, the basic VAE fails to build a direct connection between the output of the encoder and the input of the decoder, which harms the feedback accuracy. We want the output of the encoder to be fed directly into the decoder. Therefore, we reference the solution in \cite{SWAE}, where the similarity between the distribution of latent code $\mathcal{S}$ and the pre-defined distribution $\mathcal{Z}$ is measured with Wasserstein distance which can be expressed by
\begin{equation}
    W_c(p_{\mathcal{S}},p_{\mathcal{Z}})=\inf_{\gamma\in \Gamma(s\sim\mathcal{S},z\sim\mathcal{Z})}\int_{\mathcal{S}\times \mathcal{Z}}c(s,z)d\gamma(s,z),
\end{equation}
where $\Gamma(s\sim\mathcal{S},z\sim\mathcal{Z})$ is the set of all transportations plans between distributions $\mathcal{S}$ and $\mathcal{Z}$, and $c(\cdot,\cdot)$ is the distance measurement function. To numerically calculate it, a batch of latent code $\{s_m\}_{m=1}^M$ and the same size of samples from the pre-defined distribution $\{z_m\}_{m=1}^M$ are projected on a series of direction vectors $\{\theta_l\}_{l=1}^L$ sampled from a unit sphere space and are sorted in an ascending order to form sliced scalar samples sets $\{\tilde{s}^l_{[m]}=\theta_l\cdot s_m,\  \tilde{s}^l_{[m]}\leq \tilde{s}^l_{[m+1]}\}_{m=1}^M$ and $\{\tilde{z}^l_{[m]}=\theta_l\cdot z_m,\  \tilde{z}^l_{[m]}\leq \tilde{z}^l_{[m+1]}\}_{m=1}^M$. Finally, the Wasserstein distance can be calculated by
\begin{equation}
    \frac{1}{L\cdot M}\sum\limits_{l=1}\limits^{L}\sum\limits_{m=1}\limits^{M}c(\tilde{s}^l_{[m]}, \tilde{z}^l_{[m]}),
\end{equation}
where $c(\cdot, \cdot)$ is L-1 norm and L-2 norm. The whole process of Dig-CSI is shown in Algorithm \ref{alg}.

\begin{algorithm}[!ht]
\setstretch{1.2}
\small
\caption{The implementation process of Dig-CSI}
\label{alg}
\begin{algorithmic}[1]
\REQUIRE $N$ UEs participating in training, each possessing local dataset $\mathcal{X}_i$, local encoder $f_{en,i}(\cdot;\ \Theta_{en,i})$, and local decoder $f_{de,i}(\cdot;\ \Theta_{de,i})$. A pre-defined distribution $\mathcal{Z}$, distance measurement function $c(\cdot, \cdot)$, a set of direction vectors  $\{\theta_l\}_{l=1}^L$ , the global model $f(\cdot;\ \Theta)$.
\FOR{$i$ in $N$}
\STATE sample a batch of $\{x_m\}_{m=1}^M$ from $\mathcal{X}$
\STATE produce latent code $\{s_m=f_{en,i}(x_m;\ \Theta_{en,i})\}_{m=1}^M$
\STATE sample a batch of $\{z_m\}_{m=1}^M$ from $\mathcal{Z}$
\STATE construct $\{\tilde{s}^l_{[m]}=\theta_l\cdot s_m,\  \tilde{s}^l_{[m]}\leq \tilde{s}^l_{[m+1]}\}_{m=1}^M$ and \\ $\{\tilde{z}^l_{[m]}=\theta_l\cdot z_m,\  \tilde{z}^l_{[m]}\leq \tilde{z}^l_{[m+1]}\}_{m=1}^M$
\STATE get reconstructed samples $\{\hat{x}_m=f_{de,i}(s_m;\ \Theta_{de,i})\}_{m=1}^M$
\STATE update $\Theta_{en,i}$ and $\Theta_{de,i}$ by minimizing \\ $\frac{1}{M}\sum\limits_{m=1}\limits^{M}c(x_m, \hat{x}_m)+\frac{1}{L\cdot M}\sum\limits_{l=1}\limits^{L}\sum\limits_{m=1}\limits^{M}c(\tilde{s}^l_{[m]}, \tilde{z}^l_{[m]})$
\STATE send $f_{de,i}(\cdot;\ \Theta_{de,i})$ to the server as distributed generator
\ENDFOR
\STATE at the server side
\FOR{$i$ in $N$}
\STATE produce distributed fake dataset \\ $\mathcal{D}_i=\{d_{ik}=f_{de,i}(z_k;\ \Theta_{de,i})\}_{k=1}^K$ where $z_k\sim \mathcal{Z}$ 
\ENDFOR
\STATE produce global dataset $\mathcal{D}=\{\mathcal{D}_1, \cdots , \mathcal{D}_N\}$
\STATE train $f(\cdot;\ \Theta)$ with $\mathcal{D}$
\end{algorithmic}
\end{algorithm}

The structure of the encoder and the decoder of the local model is illustrated in Fig \ref{generator structure}. In the encoder, four convolution layers with $3\times3$ kernel and $stepsize=2$ are adopted to extract the feature of the input samples and reduce their dimensions. Then a fully connected layer is used to produce latent variables. The decoder which is also the generator performs contrastively as the encoder. Four transpose convolution layers with $3\times3$ kernel and $stepsize=2$ are adopted to increase the dimension. The final layer composed of a convolution layer and a tanh function fixes the range of the output variable between -1 and 1, ensuring the range of the input and the output is the same. 

\section{Experimental Results}
In this section, the results of the experiment are presented, including the setting of the scenario, dataset generation, and the performance of our system.

\subsection{Setting of Scenario and Dataset Generation}
\label{setting}
\begin{figure}[b]
    \centering
    \includegraphics[width=\linewidth] {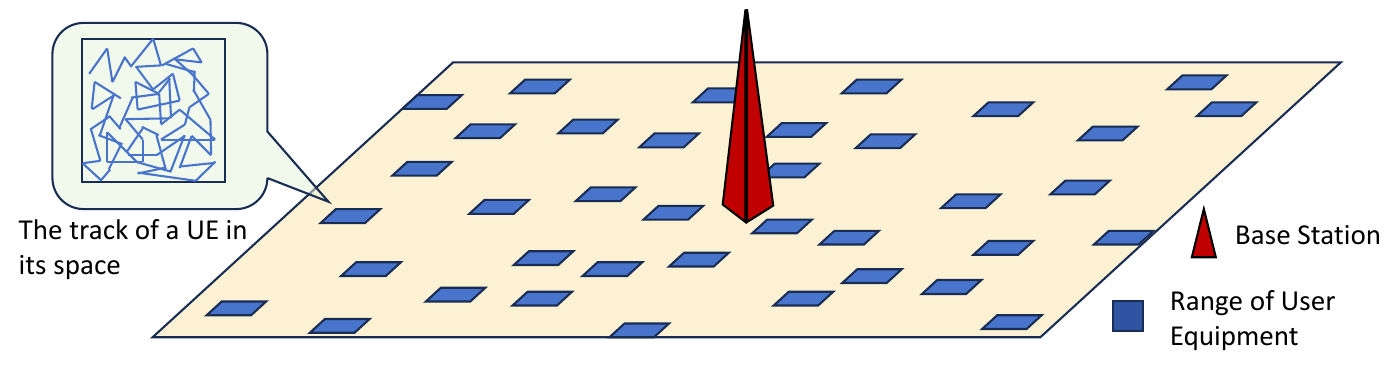}
    \caption{Scenario setting of our experiment.}
    \label{data set}
\end{figure}

To simulate the CSI training scenario with several UEs participating mentioned above, we choose QuaDRiGa \cite{qdg} for data generation. QuaDRiGa is a statistical ray-tracing wireless channel model. We define a square area with an edge length of $100m$ where a BS is in the center and 100 UEs are randomly distributed. Each UE moves randomly within a square area with an edge length of $6m$. The random-walking distance is $100m$ and a channel snapshot is taken for every $0.01m$, collecting 10,000 channel samples as its local dataset. The scenario setting is illustrated in Fig \ref{data set}. The BS is equipped with a ULA of 32 antennas operating at 2.655 GHz with a bandwidth of 70MHz and each UE is equipped with a single antenna. 32 frequency points are sampled with a uniform interval from the bandwidth. The scenario is UMi\_LOS defined in 3GPP TR 38.901 \cite{3GPP}. Each sample of channels in the angular-delay domain is normalized to a range of -1 and 1 to fit the preference of neural networks. 

\subsection{Performance and Achievements of Dig-CSI}

To explore the performance and the saving in communication overhead of Dig-CSI, we implement it in the setting mentioned in section \ref{setting} and also implement CL and FL as comparison methods. For evaluating the global CSI feedback model trained by these three methods, we build two test datasets. One is the participants test dataset $\mathcal{D}_{test}^p$ which is built by concatenating 10\% of each local dataset of the UE that participates in the training process. The other is the global test dataset $\mathcal{D}_{test}^g$ which is built by concatenating 10\% of each local dataset of all UEs. The normed mean square error (NMSE) calculated on these two datasets are named $PNMSE$ and $GNMSE$ to evaluate the performance of the trained model on the participating UEs and the whole area, respectively. They can be expressed as

\begin{align}
     PNMSE(\mathbf{H},\hat{\mathbf{H}})= {\mathbb{E}}_{\mathbf{H}\in\mathcal{D}_{test}^p}&\{\frac{\|\mathbf{H}-\hat{\mathbf{H}}\|_2^2}{\|\mathbf{H}\|_2^2}\}, \\
      GNMSE(\mathbf{H},\hat{\mathbf{H}})= {\mathbb{E}}_{\mathbf{H}\in\mathcal{D}_{test}^g}&\{\frac{\|\mathbf{H}-\hat{\mathbf{H}}\|_2^2}{\|\mathbf{H}\|_2^2}\},     
\end{align}

where $\hat{\mathbf{H}}$ is the reconstructed sample. The structure of the CSI feedback model at the server is the same as \cite{CsiNet} but other network structures can also be used.

In the CSI feedback task, the compression ratio is defined by the ratio of the dimension of latent code $s$ and the dimension of the CSI sample. We define four groups of experiments with 10, 40, 70, and 100 UEs participating in training respectively. In each group, compression ratios of 1/4, 1/8, 1/16, 1/32, and 1/64 are selected for experiments. During the experiment, we found that the current FL algorithm used for CSI feedback \cite{FedC} cannot converge in this setting, so we do not compare Dig-CSI to FL, but directly to CL which can be considered as a baseline training framework. In the implementation of CL, the number of samples that are sliced to form the global dataset is variable, simulating a situation in which different communication overhead is generated when UEs deliver local datasets to the server. 

First, we compare the communication overhead of Dig-CSI and CL under all data participation. For UEs that own 10,000 pieces of CSI matrix as local datasets participating in CSI feedback task, the communication overhead they incur is through uploading all their local datasets to the server in CL and uploading their generator trained on their local datasets in Dig-CSI. For the convenience of analysis, we consider the storage overhead of the model and dataset as their communication overhead. Table \ref{comm compare} shows the communication overheads of Dig-CSI and CL equivalently represented by storage overhead and the comparison of them.

\begin{table}[tb]
\footnotesize
\renewcommand{\arraystretch}{1}
\caption{Comparison of equivalent communication overhead of Dig-CSI and CL.}
\label{comm compare}
\centering
\begin{tabular}{m{1.1cm}<{\centering}| m{2cm}<{\centering} | m{2cm}<{\centering} | m{2cm}<{\centering}}
\specialrule{0.2em}{0pt}{2pt}
zdim &\makecell[c]{overhead of \\ Dig-CSI (KBs)} & \makecell[c]{overhead of \\ CL (KBs)} & proportion \\
\specialrule{0.2em}{2pt}{2pt}
 10&	1,609&	\multirow{7}{*}{154,510}&	1.04\% \\
 20&	1,649&	&	1.07\% \\
40&	1,729&	&	1.12\% \\
100&	1,969&	&	1.27\% \\
400&	3,169&	&	2.05\% \\
800&	4,769&	&	3.09\% \\
2000&	9,569&	&	6.19\% \\
\specialrule{0.2em}{2pt}{0pt}
\end{tabular}

\end{table}

As the storage overhead of the generator in Dig-CSI depends on the hyper-parameter of the dimension of the latent variable (marked as `zdim'), we change zdim to produce different communication overheads of Dig-CSI and list them in Table \ref{comm compare}. Although zdim encounters an increase of 200 times (from 10 to 2,000), the equivalent communication overhead (size of generator) of Dig-CSI stays relatively light compared to that of CL. We calculate the proportion of overhead of Dig-CSI in different zdims to that of CL with all local data participating in uploading and list them in the last column of Table \ref{comm compare}. Results show that in a wide range of latent variable dimensions, the overhead of Dig-CSI accounts for less than 10\% of that of CL. 

\begin{table*}[htbp]
\footnotesize
\renewcommand{\arraystretch}{1.2}
\caption{the performance of Dig-CSI and CL valued by GNMSE and PNMSE under different numbers of UE participating.}
\label{result}
\centering
\begin{tabular}{m{2cm}<{\centering} | m{2cm}<{\centering} || >{\columncolor{yellow!10}} m{1.2cm}<{\centering} |>{\columncolor{blue!5}} m{1.2cm}<{\centering} |>{\columncolor{yellow!10}} m{1.2cm}<{\centering} |>{\columncolor{blue!5}} m{1.2cm}<{\centering} |>{\columncolor{yellow!10}} m{1.2cm}<{\centering} |>{\columncolor{blue!5}} m{1.2cm}<{\centering}} 
\specialrule{0.2em}{0pt}{0pt}
\multirow{2}{*}{number of UEs} & \multirow{2}{*}{compression ratio} & \multicolumn{2}{c|}{CL} & \multicolumn{2}{c|}{Dig-CSI} & \multicolumn{2}{c}{CL (same overhead)} \\
\cline{3-8}
& & PNMSE & GNMSE & PNMSE & GNMSE & PNMSE & GNMSE \\
\specialrule{0.15em}{0pt}{0pt}
\multirow{5}*{40} & 1/4 & -17.30	&-7.02 & -14.57&-8.03	& -8.71&-4.72 \\
& 1/8 & -15.00&-6.51	& -11.91&-6.02	& -7.70&-3.53 \\
& 1/16 & -12.63&-4.82	& -9.69&-4.80	& -6.71&-2.59  \\
& 1/32 & -9.65&-4.31	& -7.86&-3.85	& -4.77&-2.18 \\
& 1/64 & -7.61&-3.69	& -5.91&-3.35	& -4.46&-1.59 \\
\specialrule{0.15em}{0pt}{0pt}
\multirow{5}*{70} & 1/4 & -17.07&-12.53	& -14.08&-12.02	& -9.13&-7.28 \\
& 1/8 &-13.66&-9.98	& -12.04&-9.58	& -6.92&-5.66 \\
& 1/16 & -11.07&-7.93	& -9.19&-7.21	& -5.72&-4.18 \\
& 1/32 & -8.44&-6.30	& -6.87&-5.44	& -3.49&-3.07 \\
& 1/64 & -6.67&-5.24	& -5.57&-4.44	& -3.60&-2.55 \\
\specialrule{0.15em}{0pt}{0pt}
\multirow{5}*{100} & 1/4 & -15.41& -15.41	& -13.49& -13.49	& -9.60& -9.60 \\
& 1/8 & -12.71& -12.71	& -11.24& -11.24	& -8.01& -8.01 \\
& 1/16 & -10.67& -10.67	& -8.83& -8.83	& -6.27& -6.27 \\
& 1/32 & -8.15& -8.15	& -6.82& -6.82	& -3.75& -3.75 \\
& 1/64 & -6.22& -6.22	& -5.04& -5.04	& -3.75& -3.75 \\
\specialrule{0.2em}{0pt}{0pt}

\end{tabular}
\end{table*}

\begin{figure}[b]
    \centering
    \includegraphics[width=\linewidth] {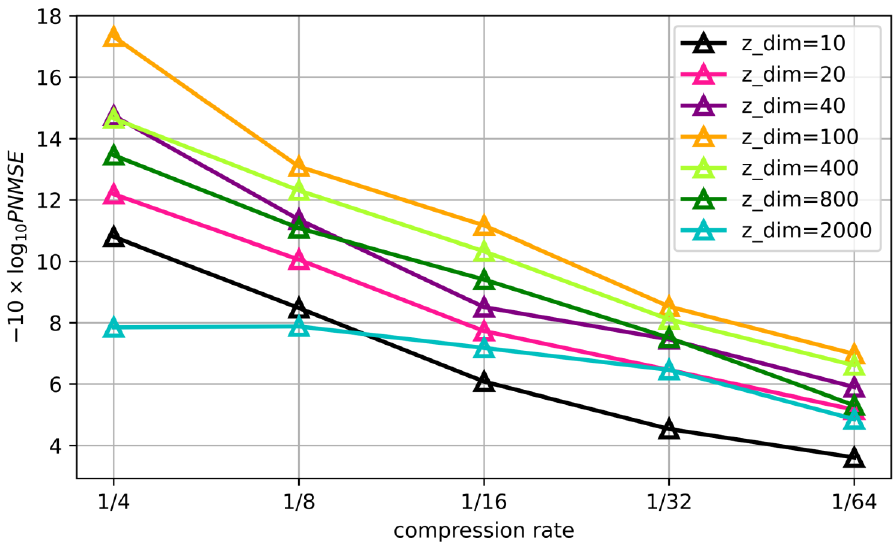}
    \caption{The performance of Dig-CSI in different dimensions of latent variables.}
    \label{zdim}
\end{figure}

In order to explore the influence of latent variable dimensions on the training of the CSI feedback model during generator training, we conduct experiments under the Dig-CSI framework using different latent variable dimension values in Table \ref{comm compare} with the participation of 10 UEs. The performance of the CSI feedback model under these experiments indicated by NMSE is presented in Figure \ref{zdim}, from which it can be observed that with the increase of latent variable dimension, the accuracy of the CSI feedback model first increases and then decreases. One possible explanation is that a too-narrow latent variable dimension fails to model the distribution of each local dataset and a too-large latent variable dimension has a space that is too complex to train. Generally speaking, it is appropriate to set the latent variable dimension equal to 100 or 400.

Next, we compare the performance of Dig-CSI and CL with the participation of 40, 70, and 100 UEs by fixing the latent variable dimension at 400. From Table \ref{comm compare}, in this condition, the overhead of Dig-CSI is 2\% of CL. We deploy two experiments for CL which are `CL (all)' where all local datasets are uploaded and `CL (same overhead)' where 2\% of local datasets are uploaded to be compared with Dig-CSI under the same training scale and the same overhead, respectively. The results are expressed in Table \ref{result} where the values of GNMSE and PNMSE are listed as dB. When considering the trained CSI feedback model serving the participating UEs which is shown in the `PNMSE' columns, the performance of Dig-CSI is slightly weaker than CL as the fake data is not exactly the true local data, but the gap is not big, demonstrating the effectiveness of Dig-CSI. When considering the trained CSI feedback model serving the whole area which is shown in the `GNMSE' columns, the performance of Dig-CSI is almost the same as that of CL, indicating that the generated fake data may slightly boost the generalization ability of the model. Furthermore, the comparison of columns `Dig-CSI' and `CL (same overhead)' shows that under the same overhead, the performance of Dig-CSI is stronger than CL, which demonstrates the advantage of Dig-CSI in a communication resources-limited scenario in the CSI feedback task.

\section{Conclusion}
This paper proposes a distributed, generative autoencoder-assisted framework for CSI feedback model training named Dig-CSI. In this framework, Each UE trains its local model implemented by an autoencoder with both feedback accuracy and generation ability and uploads the decoder as a local generator to the server for generating fake data for training a global CSI feedback model.  When training the local model, the sliced Wasserstein distance is used to make the distribution of latent code output by the local encoder close to a pre-defined distribution from which the local decoder can produce fake samples similar to local data. We simulate a scenario where several UEs with limited moving range participate in CSI feedback and compare Dig-CSI with the centralized learning (CL) method. The performance of Dig-CSI is close to CL with the same size of data participating in training, and superior with the same communication overhead.

\bibliographystyle{IEEEtran}
\bibliography{ref}

\end{document}